\documentclass[letterpaper]{article} 
\usepackage{aaai25}  
\usepackage{times}  
\usepackage{helvet}  
\usepackage{courier}  
\usepackage[hyphens]{url}  
\usepackage{graphicx} 
\usepackage{natbib}  
\usepackage{caption} 
\usepackage{algorithm}
\usepackage{algorithmic}

\usepackage{amsmath}
\usepackage{amssymb}
\usepackage{amsthm}
\usepackage{amsfonts}
\usepackage{tabularray}
\usepackage{mathtools}
\usepackage{booktabs}

\usepackage{newfloat}
\usepackage{listings}
\DeclareCaptionStyle{ruled}{labelfont=normalfont,labelsep=colon,strut=off} 
\floatstyle{ruled}
\newfloat{listing}{tb}{lst}{}
\floatname{listing}{Listing}
\title{Color Transfer with Modulated Flows}
\author{
    Maria Larchenko,
    Alexander Lobashev,
    Dmitry Guskov,
    Vladimir Vladimirovich Palyulin
}
\affiliations{
    Skolkovo Institute of Science and Technology, Moscow 121205, Russia\\

    mariia.larchenko@gmail.com, lobashevalexander@gmail.com
}

\begin{document}

\maketitle

\begin{abstract}
In this work, we introduce Modulated Flows (ModFlows), a novel approach for color transfer between images based on rectified flows. The primary goal of the color transfer is to adjust the colors of a target image to match the color distribution of a reference image. Our technique is based on optimal transport and executes color transfer as an invertible transformation within the RGB color space. The ModFlows utilizes the bijective property of flows, enabling us to introduce a common intermediate color distribution and build a dataset of rectified flows. We train an encoder on this dataset to predict the weights of a rectified model for new images. After training on a set of optimal transport plans, our approach can generate plans for new pairs of distributions without additional fine-tuning. We additionally show that the trained encoder provides an image embedding, associated only with its color style. The presented method is capable of processing 4K images and achieves the state-of-the-art performance in terms of content and style similarity.
\end{abstract}

\begin{links}
\link{Code}{https://github.com/maria-larchenko/modflows}
\end{links}

\section{Introduction}

\begin{figure}[t]
  \centering
  \centerline{\includegraphics[width=\linewidth]{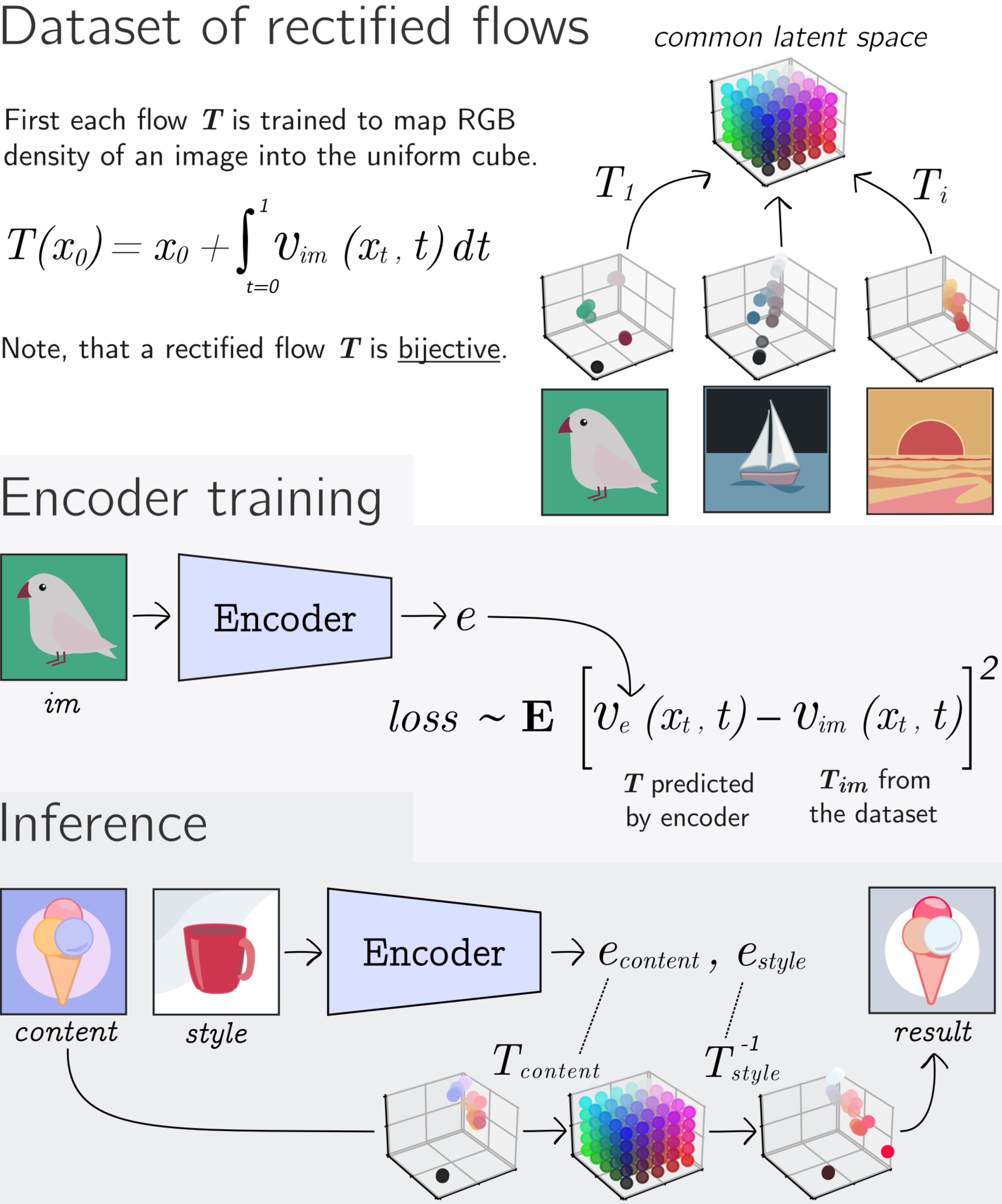}}
  \caption{A proposed scheme of training and inference. Color transfer is a composition of a forward content and an inverse style flows applied to the content image.}
  \label{fig:visual_abstract}
\end{figure}

Color adjustment is one of the most commonly used image editing operations. While minor corrections can often be made quickly, achieving a precise color palette typically requires more time and attention to details.

\textbf{Classical Methods.}
The idea to modify an image using features of another image appeared in the early 2000s \cite{jacobs2001image}.
Soon the problem of \textit{example-based color transfer} was formulated in the following way \cite{reinhard2001color}.
A pair of images known as ``content'' and ``style'' is introduced. 
The aim of the transfer is to alter the colors of the content image to fit the colors of the style image without visible distortions and artifacts.

The pioneering works on the color transfer have already considered it as a problem of optimal transport \cite{morovic2003accurate}. For instance, one would prefer to keep the shades of red as close to each other as possible. 
Technically, one defines a distance in the color space and tries to fit the desired color distribution with a minimal effort.
This effort can be seen as a transportation cost, i.e. the problem can be formulated within the framework of optimal transport (OT) theory.

In general case, an exact solution of OT problem is hard to obtain.Discretization of distributions allowed \citet{morovic2003accurate} to employ optimal histogram matching, but explicit calculation of the transport cost still was computationally heavy; for this reason other histogram-based approaches dropped the optimality constraint and considered the simpler mass preserving transport problem \cite{neumann2005color, pitie2005n}.

\begin{figure*}[h!]
  \centering
  \centerline{\includegraphics[width=\textwidth]{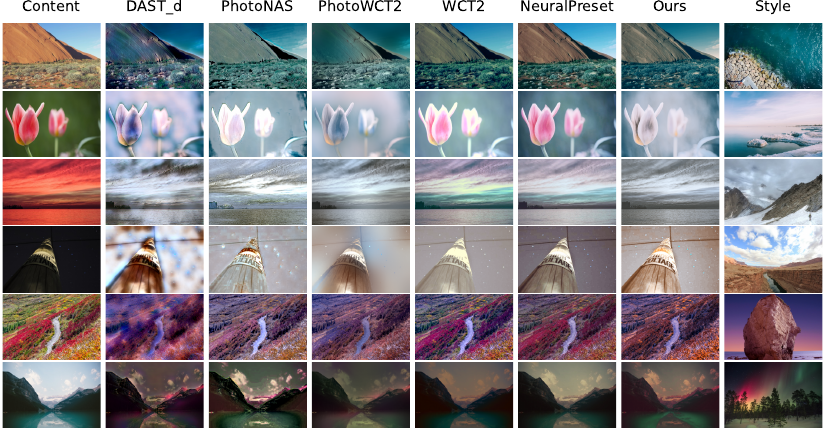}}
  \caption{Qualitative comparison. Examples from Unsplash Lite test set. Our model achieves the most exact match with the reference palette without visible distortion.}
  \label{fig:results_unsplash}
\end{figure*}

\citet{pitie2007linear} first switched to a continuous formulation of OT problem in color transfer.
Under several simplifications (e.g. that color distributions are Gaussian), authors proposed Monge-Kantorovitch Linear (MKL) algorithm, which is still a strong competitor \cite{github_mkl}.

\textbf{Neural Methods.} \citet{gatys2016image} turned the research in a different direction, adapting deep convolutional neural networks (CNNs) for a high-level style extraction.
The algorithm referred as Neural Style \cite{Johnson2015} could perform color transfer when applied to a pair of photos. However, the transfer was not ideal. It targeted a painting technique and textures, blending into stylized photos a reference color palette together with unwanted patterns.

The ability of deep CNNs to separate a color style from a content has inspired follow-up studies, primarily focusing on artifact removal. This has resulted in a series of algorithms such as
DPST \cite{luan2017deep},
WCT \cite{li2017universal},
PhotoWCT \cite{li2018closed},
WCT2 \cite{yoo2019photorealistic},
PhotoNAS \cite{an2020ultrafast},
PhotoWCT2 \cite{chiu2022photowct2},
DAST \cite{Hong_2021_ICCV}
and Deep Preset \cite{ho2021deep}. The last algorithm, aimed at automatic retouching, achieves high quality in terms of the absence of artifacts but it does not suit the color transfer task well. Nevertheless, we have included the Deep Preset in comparison to give reference scores for image retouching.

Two of the most recent studies are closely related to our work. The first one is Sparse Dictionaries \cite{huang2023optimal}, the method based on discrete optimal transport applied to learned style dictionaries. The algorithm is reported to be rather slow compared to other methods and its code is unavailable at the moment.

The second method is the Neural Preset approach proposed by \citet{ke2023neural}.
It executes \textit{color style transfer} in RGB space using a multilayer perceptron, with its hidden weights predicted by an encoder network.
It achieves impressive visual quality and is capable of processing of high-resolution images. 
The results for Neural Preset were obtained via the officially distributed application since the training code and model were not released. Due to the test set containing over a thousand images, we included Neural Preset only in the qualitative comparison.

However, we dedicated significant effort to reproducing this method. We believe that training of Neural Preset heavily depends on the random color filter adjustment strategy. In particular, the authors reported using 5000 Look-Up Table (LUT) filters, which are not publicly available. These LUT filters are designed by domain experts, and acquiring such a large number of them proved to be challenging. As a workaround, we used random monotone color filters for augmentation \cite{monotone_color_filters}. While our re-implementation successfully avoids visual artifacts, it only slightly conveys the reference visual style and has minimal impact on the color distribution. Therefore, we see the main strength of the original Neural Preset in its ability to capture the effects of hand-crafted LUT filters and we treat them as an essential part of the dataset, which could not be fully replicated by random color perturbations alone.

In order to address these limitations, we aimed on developing a model that could be trained without additional LUT filters, could be quickly applied to new images and considers the color transfer problem from the optimal transport point of view. To this end, we utilize rectified flows with parameters, predicted by an encoder network. In order to simplify the training process, we introduce a uniform latent (or intermediate) space. The rectified flows transport the color distribution of a given image to the latent space. Upon application of a particular style, we use the inverse rectified flow to transfer color distribution back from the uniform distribution to target distribution of the style image.

\textbf{Our contribution.} The contribution of this paper can be summarized as follows:

\begin{itemize}
  \item We present a novel method for color transfer based on rectified flows and a shared latent distribution. After training on a set of optimal transport plans, our approach can generate plans for new pairs of distributions without requiring additional training.
  \item We produce the dataset of 5896 flow-image pairs and train the generalizing encoder model.
  \item We show that the encoder-predicted vector of weights is an image embedding associated with its palette.
\end{itemize}

\section{Background}

\subsection{Problem setting}

In RGB space an image can be associated with a continuous 3-dimensional probability density function. We denote the density functions as $\pi_{0}$ for a content image and as $\pi_{1}$ for a style one. Here the random variables $X_0 \sim \pi_{0}$ and $X_1 \sim \pi_{1}$ represent pixels taken from the correspondent images. The color transfer problem may be defined as finding {\bf a deterministic transport map} $T(X_0) = X_1$, where $T: \mathbb{R}^D \rightarrow \mathbb{R}^D$ is a change of variables, i.e.
\begin{equation}
    \label{eq:mass_preserving_transport}
    \pi_{0}(x) = \pi_{1}(T(x)) \left|\operatorname{det} J_T(x)\right|,
\end{equation}
where $J_T(x)$ is the Jacobian of $T$ taken at point $x$.

\textbf{Monge’s optimal transportation.}
By introducing a cost function $c: \mathbb{R}^D \times \mathbb{R}^D \rightarrow \mathbb{R}$, one arrives to a minimization problem. For instance, the quadratic cost function $c(x, y) = \left\|x - y\right\|^2 $ gives a total expected cost of a transport map $T$
\begin{align}
    \label{eq:monge}
    \operatorname{Cost}\left[ T \right] = \mathbb{E}\big( \left\|X_{1} - X_{0}\right\|^{2} \big) 
    = \int_{\mathcal{X}_0} (T(x) - x)^{2} \pi_{0}(x) dx.
\end{align}

Finding of the optimal deterministic map $T^*$ that minimizes the $\operatorname{Cost}\left[ T \right]$ for a fixed cost function is called Monge problem. It does not always have a solution. However, the quadratic cost function and the continuous density functions $\pi_{0}, \pi_{1}$ with finite second moments guarantee that a solution always exists and it is unique \cite{villani2009optimal}. In some cases $T$ can be obtained explicitly. For monochrome images $X_0, X_1 \in \mathbb{R}$ and monotonically increasing cumulative distribution functions $F_0$, $F_1$ the optimal transport map $T(x)$ reads
\begin{equation}
    \label{eq:one_dimentional}
    T(x) = F_1^{-1}\big( F_0(x) \big).
\end{equation}

In practice it is possible to construct $T(x)$ even when $F_1$, $F_2$ do not have an inverse \cite{neumann2005color}. Below we make the use of this fact by proposing a new content metric, a normalized gray-scale image.

Another important case for a known $T^*: \mathbb{R}^D \rightarrow \mathbb{R}^D$ is matching of two multivariate Gaussian distributions. 
Mentioned earlier MKL by \citet{pitie2007linear} relies on the Gaussian approximations and this result.

\textbf{Monge–Kantorovich formulation.}
A correspondence between $X_0 \sim \pi_{0}$ and $X_1 \sim \pi_{1}$ can be non-deterministic. Instead of transport mapping $T$ one could consider a \textbf{transport plan} $\pi(X_0, X_1)$ (also called a coupling), a joint probability distribution with marginals $\pi_{0}$ and $\pi_{1}$,
\begin{align}
\int_{\mathcal{X}_0} \pi(x, y) dx = \pi_{1}(y),   \quad   \int_{\mathcal{X}_1} \pi(x, y) dy = \pi_{0}(x).
\end{align}


An example of a transport plan that always exists is a trivial coupling $\pi = \pi_0 \times \pi_1$, a plan where initial and target random variables are independent.

Monge–Kantorovich problem is to find $\pi^*(X_0, X_1)$ that minimizes the expected cost
\begin{align}
    \operatorname{Cost}\left[ \pi \right] = \mathbb{E}\big(  c(X_{0}, X_{1}) \big)  
    = \int_{\mathcal{X}_0\times\mathcal{X}_1} c(x, y) \pi(x, y) dx dy.
\end{align}
Let $\Pi(\pi_0, \pi_1)$ be all possible couplings of $\pi_0$ and $\pi_1$. Then the \textbf{optimal transport cost} between the initial and target distributions is
\begin{align} 
    \label{eq:optimal_cost}
    \operatorname{C}(\pi_0, \pi_1) &= \inf_{\pi \in \Pi(\pi_0, \pi_1)} \int_{\mathcal{X}_0\times\mathcal{X}_1} c(x, y) d \pi(x, y).
\end{align}
The optimal transport cost is tightly connected with the \textbf{Wasserstein distance} between two distributions. Note that the equation above is written for an unspecified cost function, i.e. the axioms of distance are not satisfied. By replacing a cost $c(x, y)$ with a proper distance function $d(x, y)$ (the quadratic cost suits this purpose) one gets a Wasserstein distance of order one
\begin{align}
    \label{eq:wasserstein}
    \operatorname{W}(\pi_0, \pi_1) &= \inf_{\pi \in \Pi(\pi_0, \pi_1)} \int_{\mathcal{X}_0\times\mathcal{X}_1} d(x, y) d \pi(x, y).
\end{align}

 \subsection{Rectified flows}
The optimal transport problem can be approximately solved by Rectified flows \cite{liu2022flow}. Its key idea is in converting an arbitrary initial coupling into a deterministic transport plan. The new transport plan guarantees to yield no larger transport cost than initial one simultaneously for all convex cost functions. First, the independent pairs \( (X_{0}, X_{1}) \) from the trivial transport plan are sampled
\begin{equation}
  \pi_{\text{trivial}}(X_{0}, X_{1}) = \pi_{0}(X_{0}) \times \pi_{1}(X_{1}).
\end{equation}
Secondly, a linear interpolation between the initial and target samples is introduced by setting\( X_{t} = t X_{1} + (1-t) X_{0} \). With this, one trains a neural network \( v_\theta(X_t, t) \) to minimize the loss
\begin{equation}
    \label{eq:ode}
    \min_{\theta} \int_{t=0}^{1} \mathbb{E}_{(X_0, X_1) \sim \pi_{\text{trivial}}} \left[ \left\| X_1 - X_0 - v_\theta(X_t, t) \right\|^2 \right ] \mathrm{d}t.
\end{equation}
Given a trained rectified flow one can transport samples from the initial distribution \( \pi_{0} \) to the target distribution \( \pi_{1} \) in a deterministic way by numerically solving the ordinary differential equation (ODE)
\begin{equation}
   \frac{d Z_t}{dt} = v_{\theta}(Z_t, t)
\end{equation} 
for \( t \in [0,1] \) with \( Z_{0} \sim \pi_{0} \). Thus, for this particular case the deterministic transport map reads 
\begin{equation}
   T_{\text{1-rectified}}(Z_{0}) = Z_{0} + \int_{t=0}^{1} v_{\theta}(Z_t, t) dt.
\end{equation} 
The deterministic transport map $T_{\text{1-rectified}}$ gives rise to the deterministic transport plan $\pi_{\text{1-rectified}}$,
\begin{equation}
   \pi_{\text{1-rectified}}(X_{0}, X_{1}) = \pi_{0}(X_{0}) \times \delta (X_{1} - T_{\text{1-rectified}}(X_{0})).
\end{equation}
This transport plan has a much lower transport cost than the na\"{i}ve transport plan $\pi_{\text{trivial}}(X_{0}, X_{1})$.

\section{Method}
Our method is inspired by the increasing rearrangement coupling \cite{villani2009optimal} given by Eq. \ref{eq:one_dimentional}.
The transfer task is complicated as we want the model to generalize well across all possible pairs $(\pi_i, \pi_j)$ of color distributions.
However, having the opportunity to learn bijective mappings, one could greatly simplify the task by introducing a common intermediate distribution $U$.

\begin{algorithm}[t]
  \caption{Encoder training} \label{alg:training}
  \small
  \textbf{Require}: trained image-flow pairs ($\mathcal{I}$, $\theta$)
  \begin{algorithmic}[1]
    \REPEAT
      \STATE get batch $\ \boldsymbol{\mathcal{I}}=\{\mathcal{I}\}_i^N, \ \ \boldsymbol{\theta}=\{\mathcal{\theta}\}_i^N$
      \FOR{$i = 1, \dotsc N$}
          \STATE  sample $X \sim \mathcal{I}$
          \STATE  $Z = \operatorname{T}_{\theta}(X)$
          \STATE  collect $t \sim \operatorname{Uniform}\left[0, 1\right]$
          \STATE  collect $Z_t = t Z + (1 - t) X$
          \STATE  collect $v_t = v_{\theta}(Z_t, t)$
      \ENDFOR 
      \STATE  Randomly reflect and rotate $\mathcal{I} \in \boldsymbol{\mathcal{I}}$
      \STATE  $\boldsymbol{e} = \operatorname{Enc}(\boldsymbol{\mathcal{I}})$
      \STATE  $\boldsymbol{t} = \{t\}_i^N, \ \ \boldsymbol{Z_t}=\{Z_t\}_i^N, \ \ \boldsymbol{v_t} = \{v_t\}_i^N$ 
      \STATE  Apply $\boldsymbol{e}$ as parameters for $\operatorname{ModFlow}$ to get $\boldsymbol{v_e}(\boldsymbol{Z_t}, \boldsymbol{t})$
      \STATE  Take gradient step with respect to $\operatorname{Enc}$ weights on $\nabla \ \mathbb{E} \left[ \left\|\boldsymbol{v_t} - \boldsymbol{v_e}(\boldsymbol{Z_t}, \boldsymbol{t})\right\|^2 \right]$
    \UNTIL{converged}
  \end{algorithmic}
\end{algorithm}

The distribution $U$ is implicitly present in the increasing rearrangement, such that for any random variable $X \sim \pi, X\in\mathbb{R}$ having monotonically increasing CDF
\begin{equation}
\begin{split}
   F(x) = \int_{-\infty}^x d\pi(y) \quad \text{it holds that} \\
   \quad U = F(X)\sim\operatorname{Uniform}\left[0, 1\right].
\end{split} 
\end{equation} 

Therefore, for a pair of such random variables $X_i, X_j \in\mathbb{R}$ a composition $T = F_j^{-1} \circ F_i$ is a transport plan that traverses through a $\operatorname{Uniform}\left[0, 1\right]$ distribution. 

We are extending this idea to random variables $X_i\in\mathbb{R}^D$ by learning bijective mappings $T_i: \mathbb{R}^D \rightarrow \mathbb{R}^D$ such that $T_i(X_i) = U^D$, where $U^D$ is random variable in $\mathbb{R}^D$ with all components uniformly distributed in $\left[0, 1\right]$. For any pair $X_i, X_j$ we define $T(X_i) = X_j$ as $T = T_j^{-1} \circ T_i$

Here rectified flow offers three important benefits. Firstly, as a solution of ordinary differential equation \ref{eq:ode} it is bijective. Secondly, it keeps the marginal distributions close to the desired ones. Lastly, the rectification step allows us to substantially increase the inference speed without adding the transport cost. Thus, we are able to efficiently compute $T$ as a composition.

During the experiments we observed that lightweight shallow models with a number of trained parameters ranging from approximately 500 to 10,000 could work as color transfer flows. The number of parameters lies in the same range with an output vector length of encoders so one may hope to use the output vector as flow parametrization, thus generalizing the approach. 

The proposed method consists of two stages:
\begin{enumerate}
    \item Produce a dataset of flow-image pairs, where flows' weights $\theta_i$ are trained to map a color distribution $X_i$ of an image $\mathcal{I}_{i}$  into the uniform cube $U$.
    We follow \cite{liu2022flow} with an interpolation $X_t = t\ U + (1-t) X_i$
    \begin{equation}
        \label{eq:flow_loss}
        \min_{\theta_i} \int_{t=0}^{1} \mathbb{E}_{(U, X_i) \sim \pi_{\text{trivial}}} \left[ \left\| U - X_i - v_{\theta_i}(X_t, t) \right\|^2 \right ] dt.
    \end{equation}
    
    \item Train the encoder on batches from the dataset, such that the output vector 
     $\operatorname{Enc}(\mathcal{I}_{i}) = e_i$ is a flow parametrization for an image $\mathcal{I}_{i}$.
\end{enumerate}
Note, that the second stage does not include any distances $d(\theta, e)$. A flow parameterized by the encoder (or the \textbf{modulated flow}) is not obliged to have the same architecture as models in a dataset. We train the encoder using the loss function that allows a distillation
\begin{equation}
    \label{eq:enc_loss}
        \min_{\operatorname{Enc}} \int_{t=0}^{1} \mathbb{E}_{(Z_i, X_i) \sim \pi_{\text{1-rectified}}} \left[ \left\| Z_i - X_i - v_{e_i}(Z_{it}, t) \right\|^2 \right ] dt,
\end{equation}
where $\operatorname{Enc}(\mathcal{I}_{i}) = e_i$ and target $Z_i$ is generated from a $X_i$ by trained flow $\theta_i$
\begin{equation}
       Z_i = T_{\text{1-rectified}}(X_i) = X_i + \int_{t=0}^{1} v_{\theta_i}(Z_t, t) dt
\end{equation}
and $Z_{it}$ are points sampled from an interpolation line connecting original $X_i$ with its target $Z_i$ 
\begin{equation}
       Z_{it} = tZ_i + (1-t)X_i.
\end{equation}
The predicted velocity $v_{e_i}(\cdot, t)$ is given by the modulated flow with $e_i$ weights. Generally, it is not advised to take the dimension of $e$ much higher than the bottleneck of selected encoder. 

Algorithm \ref{alg:training} provides the pseudo-code for the proposed method of training modulated flows. The term ``modulated'' refers to the fact that the weights of a flow at inference are produced (or modulated) by the encoder. It is important to note that the original rectified flow approach requires re-training for each new pair of densities, whereas modulation eliminates the need for this process. As demonstrated in the ablation study, using a generalizing model such as the encoder slightly increases the Wasserstein distance from the target distribution. However, it also provides implicit regularization, reducing the average Lipschitz constant of the modulated flows compared to rectified flows trained from scratch (see Table \ref{tab:lipschitz_constants}), which results in fewer visual artifacts.

\begin{center}
\begin{table*}
\begin{minipage}[t]{0.45\textwidth}
      \centering
      \begin{tabular}{llll}
        \toprule
        \multicolumn{2}{c}{Aggregated scores (DISTS)$\downarrow$} \\
        \cmidrule(r){1-2}
        Algorithm       & Grayscale  & Depth &  Edge \cite{xie15hed} \\
        \midrule
        ModFlows (ours)  & \bf 0.129 	 & \bf 0.217 	 &  \underline{0.220} \\
        MKL 	 	 & \underline{0.146} &  \underline{0.227} 	 & 0.224 \\
        CT 	 	     & 0.169 	 &  0.234 	 & 0.232 \\
        WCT2 	 	 & 0.170 	 &  0.228 	 & 0.249 \\
        PhotoWCT2 	 & 0.191 	 &  0.236 	 & \bf 0.217 \\
        DAST\_d (vanilla) 	 & 0.204 	 &  0.267 	 & 0.224 \\
        DAST\_da (adversarial)	 & 0.214 	 &  0.282 	 & 0.229 \\
        PhotoNAS 	 & 0.224 	 &  0.276 	 & 0.270 \\
        NeuralPreset* 	 & 0.349 	 &  0.366	 & 0.360 \\
        Deep Preset 	 & 0.384 	 &  0.400 	 & 0.387 \\
        \bottomrule
      \end{tabular}
\end{minipage}
\hfill
\begin{minipage}[t]{0.45\textwidth}
      \centering
      \begin{tabular}{ll}
        \toprule
        \multicolumn{2}{c}{Style distance$\downarrow$} \\
        \cmidrule(r){1-2}
        Algorithm        & mean $\pm$ std of mean\\
        \midrule
        DAST\_d & 	 \textbf{0.112 $\pm$ 0.001}\\
        ModFlows (ours)& \underline{0.123 $\pm$ 0.001}\\ 
        DAST\_da & 	 0.127 $\pm$ 0.001\\
        PhotoWCT2 & 	 0.129 $\pm$ 0.001\\
        MKL & 	 0.145 $\pm$ 0.001\\
        WCT2 & 	 0.163 $\pm$ 0.001\\
        CT & 	 0.166 $\pm$ 0.001\\
        PhotoNAS & 	  0.183 $\pm$ 0.002\\
        NeuralPreset* 	 & 0.348 $\pm$ 0.003 \\
        Deep Preset & 0.384 $\pm$ 0.004\\
        \bottomrule
      \end{tabular}
\end{minipage}
\caption{Comparison of algorithms. Please note that NeuralPreset* is our re-implementation.}
\label{tab:aggregated_scores}
\end{table*}

\begin{figure*}
  \centering
  \centerline{\includegraphics[width=\textwidth]{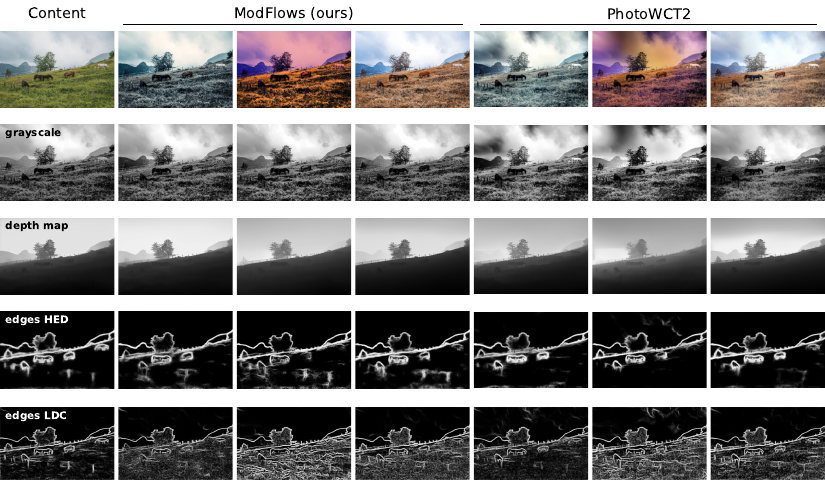}}
  \caption{Colorless content metrics. The choice of the best content metric is not obvious. Edges detection by HED model \cite{xie15hed} grasps mostly the main objects of a scene, while canny LDC \cite{soria2022ldc} images are capturing the too detailed edges. Both of them are not sensitive to low-frequency artifacts. To show the absence of such artifacts in the Modflows we additionally compute similarity scores between the normalized grayscale images, which are processed to have a linear intensity histogram through histogram matching, and the depth maps \cite{gui2024depthfm}.}
  \label{fig:colorless_metrics}
\end{figure*}

\begin{table*}
      \centering
      \begin{tabular}{llll|l}
        \toprule
        \multicolumn{2}{c}{Aggregated ablation scores (DISTS)$\downarrow$} \\
        \cmidrule(r){1-4} 
        Algorithm       & Grayscale  & Depth \cite{gui2024depthfm} &  Edge \cite{xie15hed} &  Style Distance$\downarrow$\\ 
        \midrule
        ModFlows (B6), $dim(e)$ = 8195 &\bf 0.129 &\bf 0.217 &\bf 0.220 	 & 0.123\\
        Rectified flows (8195)         &   0.137  &    0.250 &    0.235 	 & \bf 0.114\\
        ModFlows (B0), $dim(e)$ = 515  &   0.145  &\bf 0.217 &\bf 0.220 	 & 0.141\\
        \bottomrule
      \end{tabular}
      \caption{Ablation study}
      \label{tab:ablations-table}
\end{table*}
\end{center}

\section{Experiments and Metrics}

\textbf{Dataset. } 
To implement the approach described above one needs a dataset of images with sufficiently diverse color distributions and resolutions. To achieve this diversity we construct our dataset by combining DIV2K \cite{Ignatov_2018_ECCV_Workshops} and CLIC2020 \cite{CLIC2020} (designed for image compression challenges) with a subset of ``laion-art-en-colorcanny'' \cite{laion_art_en_colorcanny}. The total number of images is 5,826.

For every image we train a small two-layer MLP with 1024 hidden units (8195 parameters in total) and tanh activation, storing in the dataset 5,826 rectified models. Generation of a model-image pair takes approximately 100k iterations with lr = 5e-4.

\textbf{Encoder.} 
EfficientNet B6 is used as an encoder model \cite{tan2019efficientnet}. For simplicity we set the output dimension to 8195 for it to be the same with the dataset of trained flows. The encoder was trained with Adam optimiser \cite{kingma2014adam} for 751k iterations with the batch size equals to 8 images. We decreased the learning rate from lr = 5e-4 to lr = 1e-4 after the first 100k iterations. 

\textbf{Test set. }
Tests were conducted on 1891 content-style pairs selected from Unsplash Lite 1.2.2 \cite{unsplash_lite}. Searches were run on 25,000 Unsplash pictures. Our pictures are generated in 8 steps of ODE solver (16 steps in total for forward and inverse passes).

\textbf{Style metric. }
The seminal work \cite{gatys2016image} defines style loss as a distance between Gram matrices of feature maps, taken from convolutional layers of VGG encoder.
Despite being capable of extracting a palette, this approach cannot reliably separate a palette from textures. Monge's problem (Eqs. \ref{eq:mass_preserving_transport} and \ref{eq:monge}) offers a more precise setting and a straightforward metric, namely, Wasserstein distance, Eq. \ref{eq:wasserstein}. Therefore we estimate the Wasserstein distance between resulting and reference color distributions taking 6,000 pixel samples for a style metric \cite{bonneel2011displacement,flamary2021pot}.

\textbf{Content metric. } Contrary to the style, a content metric is not uniquely defined. To measure the amount of visible artifacts we compute a set of colorless metrics based on depth-maps by recently released DepthFM \cite{gui2024depthfm}, normalized grayscale pictures and edge-maps by HED  \cite{xie15hed, pytorch-hed} and LDC \cite{soria2022ldc} models. The variants of the colorless representation are demonstrated in Fig. \ref{fig:colorless_metrics}. The difference between colorless images is evaluated with DISTS\footnote{DISTS implementation is taken from ``piq'' library \cite{piq}} \cite{ding2020DISTS} producing the content score. 

\textbf{Lipschitz constant. } To estimate the regularity of learned color transfer maps we estimate their average Lipschitz constant, Table \ref{tab:lipschitz_constants}. It could be observed that rectified flow trained from scratch for a given pair of color distributions is more sensitive to input variations than, for instance, MKL and CT, meaning higher amount of visual artifacts. Low value of the Lipschitz constant for ModFlows encoder in comparison to the direct flows demonstrate regularizing effect of our training procedure.

\begin{table}[ht]
\centering
\begin{tabular}{|l|c|}
\hline
\textbf{Method} & \textbf{Average Lipschitz Constant} \\
\hline
ModFlows (ours) & \bf 37.26 $\pm$ \bf 0.79 \\
CT & 51.90 $\pm$ 1.03 \\
MKL & 55.67 $\pm$ 1.19 \\
Rectified flows & 91.34 $\pm$ 1.26 \\
DAST\_d & 121.17 $\pm$ 1.76 \\
PhotoWCT2 & 160.12 $\pm$ 1.92 \\
\hline
\end{tabular}
\caption{Average Lipschitz constant of the color transfer map for different methods. Low value of the Lipschitz constant for ModFlows encoder in comparison to the direct flows demonstrate regularizing effect of our training procedure.}
\label{tab:lipschitz_constants}
\end{table}

\textbf{Comparison with baselines. } Table \ref{tab:aggregated_scores} contains average style distances and aggregated scores for compared methods. 
Please note that NeuralPreset* is our re-implementation of the original work by \citet{ke2023neural}. It was trained on the same dataset as our method, but the LUT filters were replaced with random color perturbations \cite{monotone_color_filters} since the original color filters and model are not available.

The \textbf{aggregated score} is calculated as a distance to the ideal point $p$, similarly with \citet{ke2023neural},
\begin{equation}
\begin{split}
\text{aggr. score}= \sqrt{(\text{p} - \text{style score})^2 + (\text{p} - \text{content score})^2}
\end{split}
\end{equation}

\begin{center}
    \begin{figure}[t]
        \includegraphics[width=\linewidth]{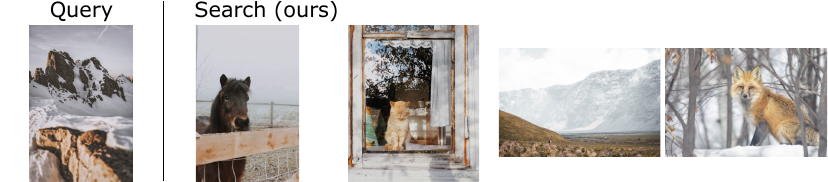}
        \includegraphics[width=\linewidth]{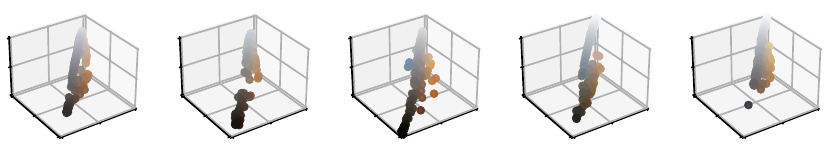}
        \includegraphics[width=0.6\linewidth]{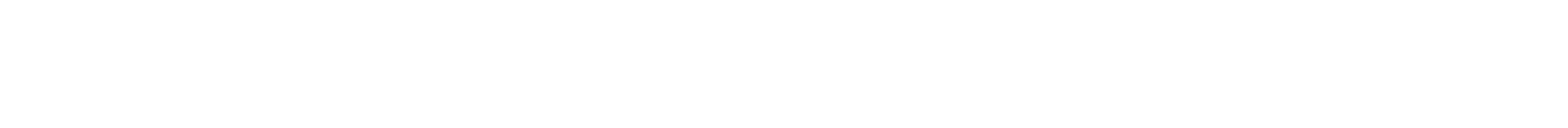}
        \includegraphics[width=\linewidth]{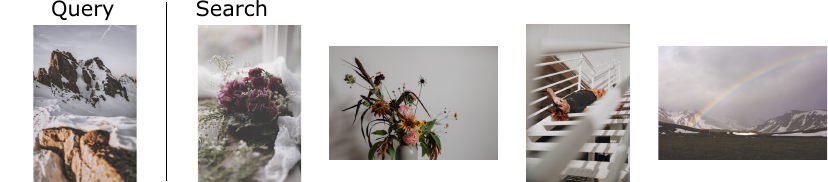}
        \includegraphics[width=\linewidth]{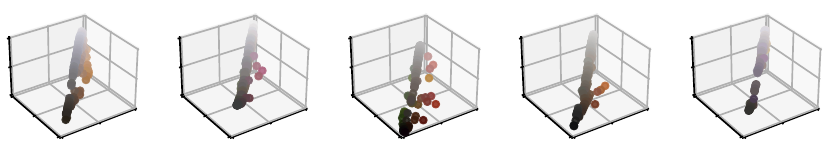}
        \caption{Search for similar color styles in the Unsplash Lite dataset (25k images). The top and second rows show search results based on the output of the ModFlows (B6) model. The third and last rows display results based on image statistics, specifically flattened vectors representing the first and second centered moments of the color distribution.}
        \label{fig:search_ours}
    \end{figure}
\end{center}

\textbf{Search of similar color styles. } Once trained, the output vector of parameters $e$ could serve as an embedding of a palette. To evaluate its expressive ability we compare $e$ against standard statistics for RGB channels $(\boldsymbol{\mu}, \Sigma)$, that is, the vector of mean values concatenated with flattened covariance matrix. An example of a search is given in Fig. \ref{fig:search_ours}.

\section{Ablation Study}
\begin{figure}[h!]
  \centering
  \centerline{\includegraphics[width=\linewidth]{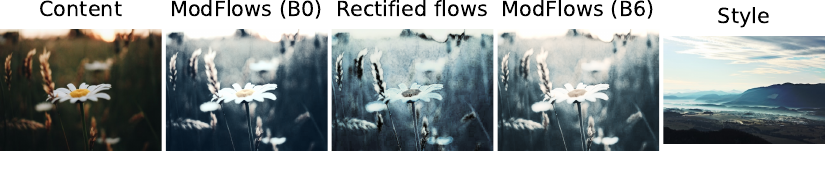}}
  \centerline{\includegraphics[width=\linewidth]{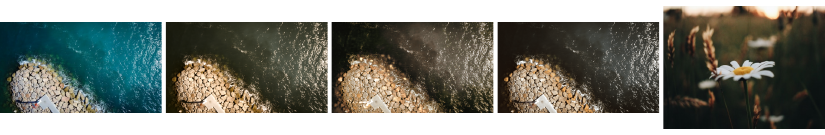}}
  \caption{Ablation study. ModFlows models reach a better trade-off between style and content similarity when compared to dataset models used in their training.}
  \label{fig:ablation}
\end{figure}

All comparisons in this section are computed on a test set described above. We describe qualitatively and numerically the performance of
\begin{enumerate}
    \item Transfers made with rectified flows (8195 parameters) 
          \newline through the uniform intermediate space.
    \item Model based on EfficientNet B6 with output $dim(e)$ = 8195
          trained on 5,826 flows-8195 from the main dataset.
    \item Model based on EfficientNet B0 with output $dim(e)$ = 515
          trained on 4,767 rectified flows (515 parameters) from the laion-art-en-colorcanny.
\end{enumerate}

As the Table \ref{tab:ablations-table} proves, the low style distance in transfers made with rectified flows comes with artifacts which are detected by all content metrics, which is shown in Fig. \ref{fig:ablation}. At the same time the generalization done by the ModFlows models reaches a better trade-off between style and content similarity. As expected, providing larger and more diverse dataset along with increased number of parameters results in a better performance.

From our experiments it follows that choosing another color space such as LAB or OkLAB \cite{ruderman1998statistics} doesn't significantly improve the results. Despite these spaces offers better perceptual distance, they additionally complicate a training procedure, namely, a shape of a suitable shared latent space and the sampling process.

\section{Limitations and Algorithm Tuning}
\begin{figure}[h]
  \centering
  \includegraphics[width=\linewidth]{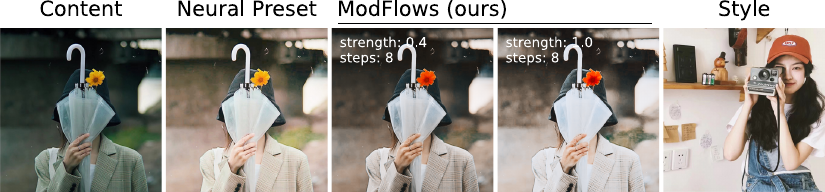}
  \includegraphics[width=\linewidth]{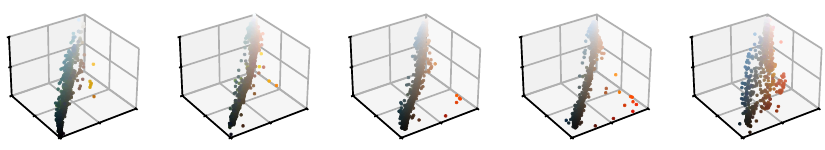}
  \caption{Limitations and algorithm tuning. An example of unintended color switching in two pictures generated with fixed number of steps for ODE solver (steps) and varied percent of interpolation curve passed (strength).}
  \label{fig:color_replacement}
\end{figure}

\begin{figure}[h]
  \centering
  \includegraphics[width=\linewidth]{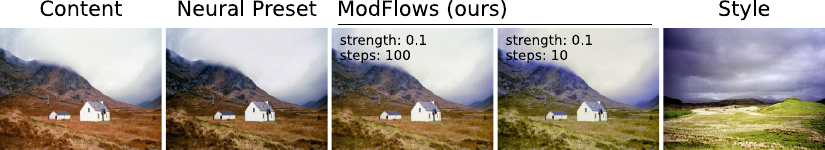}
  \includegraphics[width=\linewidth]{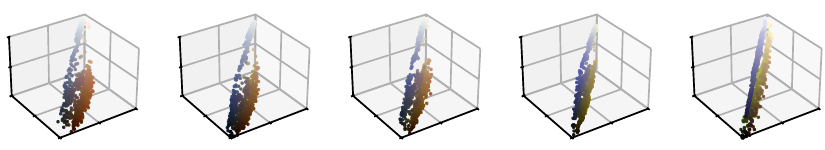}
  \caption{Algorithm tuning. Variation of a number of steps for ODE solver (steps) and a percent of interpolation curve passed (strength) results in different amount of changes for a distribution. In this example, increasing the strength or decreasing the number of steps further leads to the appearance of artifacts.}
  \label{fig:scotland_house_imgs}
\end{figure}

The framework of the transport theory gives us an opportunity to design an unsupervised algorithm. In the same time it introduces a limitation, that is a greater dependence of the result on the reference image. For example, the method may perform unintended color replacements, such as transforming yellow shades into red ones, Fig. \ref{fig:color_replacement}. Our method does not provide control over the color of individual objects, as it operates in RGB space without considering semantic information.

The presented model is able to change a color distribution significantly. Hence, in some cases the strength of transformation should be controlled to avoid artifacts and to achieve a satisfying result. In addition to a linear interpolation between original and resulting image, in a rectified flow model there are two parameters of generation process that naturally control the strength of transfer, namely, a number of steps for ODE solver and a percent of interpolation curve passed (strength) after which generation is stopped. The transfer examples where these two parameters are varied are given in Figs. \ref{fig:color_replacement} and \ref{fig:scotland_house_imgs}. 

\section{Conclusion}

We have introduced a novel approach to color transfer, a process that modifies the colors of an image to match a reference palette, such as the color distribution of a style image. Trained on a set of unlabeled images with diverse color styles, our transfer model offers a unique method of performing color transfer as a density transformation in RGB color space. The use of rectified neural ODEs to learn mappings between color distributions is a significant departure from existing methods. The existence of an inverse function of the ODE allows us to introduce a common latent space for all densities. By constructing a transformation as a composition of a forward and an inverse pass through the latent space, we simplifying the training of generalizing model, which is able to predict the mappings for new content-style image pairs.

The proposed approach outperforms existing state-of-the-art neural methods for color transfer. Furthermore, it is not restricted to a specific domain and can be applied to other areas where an image is associated with a distribution, and distribution transfer is needed.

\bibliography{main}

\end{document}